\documentclass[conference]{IEEEtran}
\IEEEoverridecommandlockouts
\usepackage{amssymb}
\usepackage{graphicx}
\usepackage{caption}
\usepackage{subcaption}
\usepackage[bookmarksnumbered,unicode]{hyperref}
\usepackage{array}
\usepackage{siunitx}
\usepackage{amsmath}
\usepackage{microtype}
\usepackage{etoolbox}
\usepackage{booktabs}
\usepackage{refcount}
\usepackage{totpages}
\usepackage{environ}
\usepackage{setspace}
\usepackage{textcase}
\usepackage{tabularx}
\usepackage{multirow}
\usepackage{xcolor}
\usepackage{cellspace}
\usepackage{booktabs}
\usepackage{caption}
\usepackage{multicol}
\usepackage{enumitem}
\usepackage{authblk}

\usepackage{geometry}
\usepackage[noline, ruled,noend,nofillcomment]{algorithm2e} 
\DontPrintSemicolon

\definecolor{Game Context}{RGB}{0,100,0}
\definecolor{GPT-4(V) Output (Facial Description)}{RGB}{0,0,100}
\SetAlFnt{\small}
\SetAlCapFnt{\small}
\SetAlCapNameFnt{\small}
\SetAlCapHSkip{0pt}
\IncMargin{-\parindent}

\usepackage{microtype}

\newcommand{\annotate}[2]{%
  \begingroup
  \setbox0=\hbox{\footnotesize#2}%
  \rlap{\raisebox{3ex}{\hbox to \wd0{\hfil\color{#1}\scriptsize#1\hfil}}}%
  $\overbrace{\hbox{\usebox0}}$%
  \endgroup
}

\begin{document}

\author{Youssef Mohamed$^{1}$, S\'everin Lemaignan$^{2}$, Arzu G\"uneysu$^{3}$, Patric Jensfelt$^{1}$, and Christian Smith$^{1}$%
\thanks{*This work was in part financially supported by Digital Futures}%
\thanks{$^{1}$Youssef Mohamed, Patric Jensfelt, and Christian Smith are with KTH: The Royal Institute of Technology, Stockholm, Sweden
        {\tt\small \{ymo, patric, ccs\}@kth.se}}%
\thanks{$^{2}$S\'everin Lemaignan is with PAL Robotics, Barcelona, Spain
        {\tt\small severin.lemaignan@pal-robotics.com}}%
\thanks{$^{3}$Arzu G\"uneysu is with Ume\r{a} University, Ume\r{a}, Sweden
        {\tt\small arzu.guneysu@umu.se}}%
}
\title{Fusion in Context: A Multimodal Approach to Affective State Recognition}

\maketitle

\begin{abstract}
Accurate recognition of human emotions is a crucial challenge in affective computing and human-robot interaction (HRI). Emotional states play a vital role in shaping behaviors, decisions, and social interactions. However, emotional expressions can be influenced by contextual factors, leading to misinterpretations if context is not considered. Multimodal fusion, combining modalities like facial expressions, speech, and physiological signals, has shown promise in improving affect recognition. This paper proposes a transformer-based multimodal fusion approach that leverages facial thermal data, facial action units, and textual context information for context-aware emotion recognition. We explore modality-specific encoders to learn tailored representations, which are then fused using additive fusion and processed by a shared transformer encoder to capture temporal dependencies and interactions. The proposed method is evaluated on a dataset collected from participants engaged in a tangible tabletop Pacman game designed to induce various affective states. Our results demonstrate the effectiveness of incorporating contextual information and multimodal fusion for affective state recognition.
\end{abstract}

\paragraph{Keywords} Human detection, computer vision, social human-robot interaction.

\section{Introduction}
Accurately perceiving and interpreting human emotions is a fundamental challenge in affective computing and human-robot interaction (HRI) fields. Since emotions significantly influence our behavior, decisions, and social interactions, creating systems that can reliably recognize and understand these emotional states is essential. Such advancements would enable robots to engage in more natural, effective interactions by better comprehending and responding to human needs and preferences.

One of the key challenges in affective state recognition lies in accounting for the contextual information surrounding affect expressions. In other words, the same facial expression or physiological signal may convey different affective meanings depending on the context in which it occurs. For instance, a smile in a social setting may indicate enjoyment, while a similar smile in a different context could signify sarcasm or discomfort. Failing to consider these contextual factors can lead to misinterpretations and inaccurate affect recognition, hindering the effectiveness of affective computing systems.

Multimodal fusion, which integrates diverse data streams such as facial expressions, speech, and physiological signals, has shown promise in enhancing affect recognition performance compared to unimodal methods~\cite{baltruvsaitis2018multimodal,mohamed2023multi,mohamed2022automatic,bano2018multimodal}. However, effectively fusing these modalities and incorporating contextual information remains a significant challenge in the field.

To address this challenge, we propose a transformer-based multimodal fusion approach for context-aware emotion recognition. Our method leverages recent advances in deep learning, particularly transformer architectures, which have demonstrated remarkable capabilities in capturing temporal dependencies and modeling complex interactions between modalities~\cite{vaswani2017attention,ma2021facial}.

Our approach utilizes modality-specific encoders to extract tailored representations from facial thermal data\cite{mohamed2022automatic}, action units\cite{barrett2019emotional}, and textual context information\cite{lemaignan2024social}. These representations are then combined and processed by a shared transformer encoder. This architecture enables effective integration of contextual cues and leverages the complementary information from multiple modalities, providing a more comprehensive understanding of affective states.

We evaluate our method on a dataset collected from participants in a tabletop Pacman game \cite{pacmandesign} re-designed to elicit various affective states, including enjoyment, boredom, and frustration. By doing so, we aim to demonstrate the potential of our approach in real-world emotion recognition scenarios.

The primary contributions of this paper are as follows.

\begin{itemize}
    \item \textbf{Contextual Information Integration:} Demonstrates the importance of incorporating contextual information as a separate modality to enhance affect recognition accuracy when added to other physiological and visual modalities.
    
    \item \textbf{Transformer-based Multimodal Architecture:} Proposes a transformer architecture with additive fusion of modality-specific representations, effectively capturing temporal dependencies and interactions for improved affective state detection.
\end{itemize}

Our experimental results on the Pacman game dataset demonstrate the relative effectiveness of different modalities and their combinations in our transformer-based multimodal fusion approach. By examining various configurations of thermal data, action units, and contextual information, we provide insights into the contributions of each modality to affective state recognition.

\section{Background}

\subsection{Context-Aware Emotion Recognition}

Context-aware emotion recognition is crucial for improving affective computing systems' accuracy \cite{hoang2021context,wang2019context}. Various approaches have been proposed to incorporate context, each with its own limitations.

Mittal et al.~\cite{mittal2021multimodal} developed a multimodal and context-aware model using multiplicative fusion to combine facial expressions, speech, and physiological signals. It employs a graph-based attention mechanism to weight modalities based on context. However, this approach may struggle with complex, non-linear relationships between modalities and context.

Wang et al.~\cite{wang2019context} proposed a context-aware network with a hierarchical attention mechanism for video data. It learns to focus on salient emotional cues, considering facial expressions, body language, voice tone, and environmental context simultaneously. The reliance on predefined hierarchical structures, however, may limit its adaptability to diverse scenarios.

Kim et al.~\cite{kim2016emotion} introduced a deep semantic feature fusion approach for video emotion recognition, combining facial expressions, audio features, and textual context using hierarchical fusion. While innovative, their method may not fully capture the nuanced interplay between different contextual elements and emotional expressions.

These studies demonstrate the importance of incorporating context in affect recognition systems, but often treat context as a fixed set of features, potentially overlooking its dynamic nature.

We propose a transformer-based architecture for more effective multimodal integration, overcoming limitations of fixed fusion strategies. By leveraging natural language processing and transformers, our method achieves a more comprehensive and adaptable integration of context in emotion recognition, potentially leading to more accurate and robust affective computing systems.

\subsection{Multimodal Machine Learning}

Multimodal machine learning integrates multiple modalities like text, audio, images, and videos~\cite{liang2022foundations}. Key principles driving innovations include modality heterogeneity, connections, and interactions~\cite{liang2022foundations}. These principles are fundamental to our work, integrating facial thermal data, action units, and textual context.

Core technical challenges include representation and alignment~\cite{baltruvsaitis2018challenges}. Representation involves encoding diverse modalities with distinct statistical properties, while alignment concerns mapping corresponding elements across modalities. These challenges are particularly relevant in emotion recognition, where facial expressions, thermal, and contextual information must be coherently integrated. 

Jiang et al.~\cite{jiang2023understanding} highlighted the importance of constructing meaningful latent modality structures, suggesting that exact modality alignment may not be optimal for tasks like emotion recognition.

Our approach addresses these principles through modality-specific encoders and a shared transformer encoder capturing temporal dependencies and interactions.

\subsection{Transformer Multimodal Fusion}

Transformer-based architectures have gained popularity for multimodal fusion due to their ability to capture inter-modality interactions and model temporal dependencies.

A recent advancement in this field is the work of Faye et al.~\cite{faye2024context}, who proposed the Context-Based Multimodal Fusion (CBMF) model. This approach combines modality fusion and data distribution alignment using context vectors fused with modality embeddings. The CBMF model is particularly relevant to our work as it shares our focus on integrating contextual information directly into the fusion process.

Our method builds upon recent advancements in multi-modal analysis for manipulation detection. While some approaches use complex interaction mechanisms~\cite{wang2024exploiting}, we employ separate encoding processes for each modality followed by additive fusion, allowing effective integration without intricate cross-modal attention mechanisms.

\section{Dataset}

This study utilizes a dataset collected by~\cite{mohamed2023multi}, which captures participants' affective states during a tangible Pacman game designed with multiple configurations and can induce four affects: frustration, enjoyment, boredom and neutral \cite{mohamed2023multi,pacmandesign}.

\subsection{Data Modalities}

The dataset comprises three main modalities: thermal data from facial regions of interest (ROIs), visual data in the form of Action Units (AUs) extracted from RGB images, and text data from game-play logs and settings (see Table~\ref{tab:features}).

\subsubsection{Thermal and Visual Features}

\begin{figure*}
    \centering
    \includegraphics[width=0.6\linewidth]{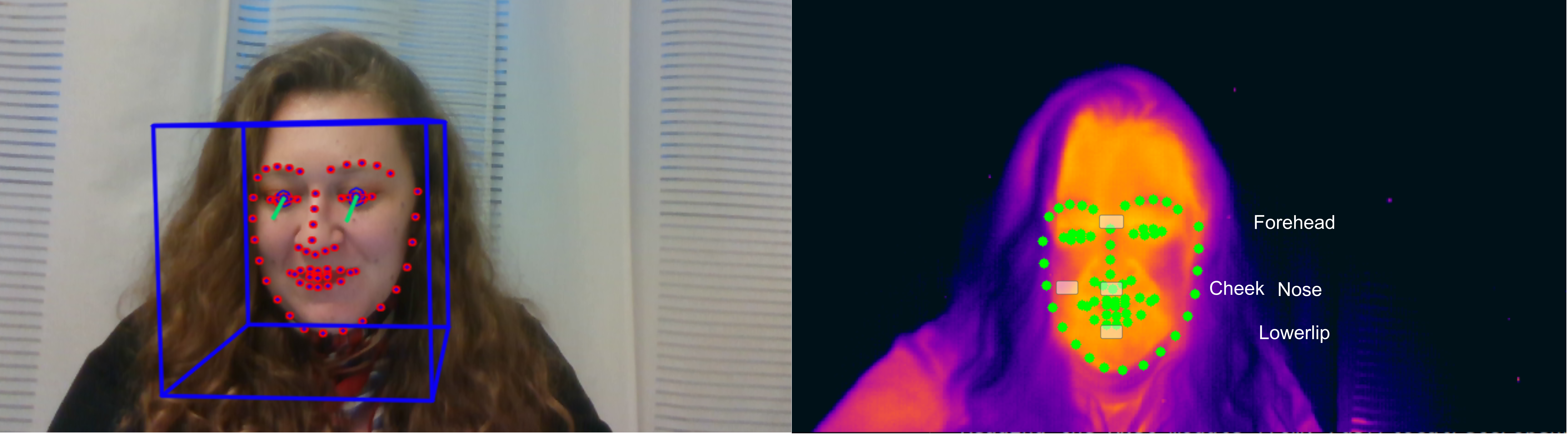}
    \caption{Facial landmarks and Action Units extracted using OpenFace (left) and thermal regions of interest (right)\cite{mohamed2023multi}.}
    \label{fig:thermalimage}
\end{figure*}

Figure~\ref{fig:thermalimage} illustrates the facial landmarks and AUs extracted using OpenFace (left), and the thermal ROIs (right).

Table~\ref{tab:features} summarizes the extracted features for each modality.

\begin{table}
\caption{Features for thermal, visual, and text modalities}
\label{tab:features}
\small
\begin{tabular}{@{}p{1.8cm}p{5.7cm}@{}}
\toprule
\textbf{Modality} & \textbf{Features} \\
\midrule
Thermal & \textbf{ROIs:} Nose, Forehead, Cheek, Lower lip\\
 & \textbf{Metrics:} Avg, Change, Max, Min temperature \\
\midrule
Visual & \textbf{AUs:} 1, 2, 4-7, 9, 10, 12, 14, 15, 17, 20, 23, 25, 26, 28, 45 \\
 & \textbf{Metrics:} Avg, Change, Max, Min intensity \\
\midrule
Text & \textbf{Game Outcomes:} Win/Loss \\
 & \textbf{Round Settings:} Fruit, Ghosts, Speed, Rotation \\
\bottomrule
\end{tabular}
\end{table}

\subsection{Dataset Composition}

Our dataset includes four distinct affective states: baseline (neutral), enjoyment, boredom, and frustration. We use a 7-second window for analysis, aligning with the methodology of \cite{mohamed2023multi}. This approach is grounded in the well-established understanding that physiological signals typically manifest on the face within a 5-15 second timeframe, while facial expressions can take up to 4 seconds to appear and often persist for longer\cite{ekman2005}.

\section{Methodology}

\subsection{Context Extraction and Classification}

We have gained access to the raw video data, the videos of each participant were then processed to capture their interactions within the game environment. The contextual data was categorized into two types: \textbf{Game-Only Context (GOC)} and \textbf{Full Context (FC)}

For both types of context, we first generated descriptive sentences, which were then transformed into high-dimensional vector representations using the OpenAI embedding model \textit{embedding-large}~\cite{openaiembeddings}. This model converted each descriptive sentence into a 3072-dimensional vector, capturing the semantic nuances of the context.

\subsubsection{Game-Only Context (GOC) Embedding}
For the GOC, we included only game-related information. An example sentence might be:

\begin{quote}
\textit{"The person is playing a Pacman game with difficulty level: easy"}
\end{quote}

The description of the game difficulty was based on the speed of the robots, their number, and the amount of rotation needed to collect the points, which was classified into three settings: easy, medium, and hard~\cite{mohamed2023multi}.

\subsubsection{Full Context (FC) Embedding}
For the FC, we combined the game settings and difficulty level with facial expression descriptions. To capture the temporal dynamics of facial expressions, we implemented a sliding window on the raw video data. We sub-sampled the video stream to 1 frame per second, as we do not expect the signals to move at a faster rate ~\cite{ekman2005}.

For each instance, we extracted a pair of consecutive frames: the current frame at time $t$ and its predecessor at time $t-1$. This two-frame window slides throughout the duration of the video, allowing us to capture the evolution of facial expressions over time~\cite{lu2024gpt}.

We used GPT-4(V) model, accessed through its API, for facial expression analysis. Each frame pair was submitted to the model using the following prompt:

\begin{quote}
\label{gptprompt}
\textit{Given two images, the first of the face at time $t-1$ and the second at time $t$, describe the current emotional state of the person in one brief sentence, considering the presence and intensities of facial expressions.}
\end{quote}

An example of a full context sentence combining game information and facial expression data would be:

\begin{quote}
\itshape\setlength{\baselineskip}{2\baselineskip}
\noindent\annotate{Game Context}{The person is playing a Pacman game with difficulty level: easy}

\noindent\annotate{GPT-4(V) Output (Facial Description)}{with a look of wonder or amazement with raised eyebrows}
\end{quote}

\subsection{Transformer}

In this section, we present a transformer-based model for multimodal affect recognition. The three inputs comprises thermal features, visual features (action units), and contextual data. The raw input is discretized using a 7-second sliding window~\cite{mohamed2022automatic} stepping 1 second at a time to match the classification into one of the four affective states provided by~\cite{mohamed2023multi}. The latter is the target output for the transformer network.

In the following subsections, we present our transformer model, discussing its architecture, key components, and the hyperparameters used for optimization.

\subsubsection{Architecture}

\begin{figure*}[t]
    \centering
    \vspace{-4em}  

    \includegraphics[width=0.8\textwidth]{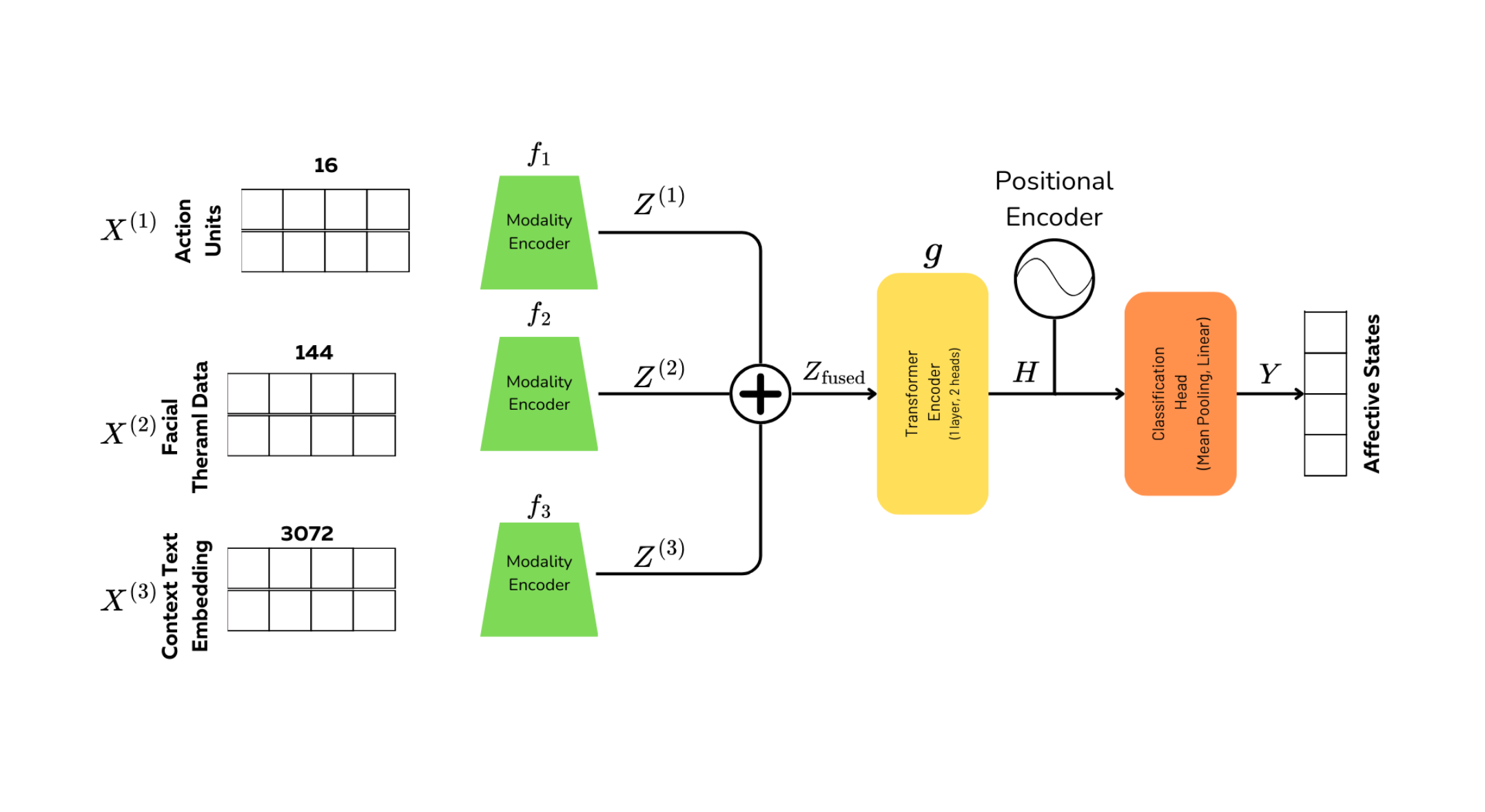}
    \captionsetup{aboveskip=-10pt, belowskip=-10pt, font=small, width=.75\textwidth}
    \caption{Multimodal Transformer Architecture: Integrates action units (16), facial thermal data (144), and context text embeddings (3072) through modality-specific encoders. Additive fusion processed by transformer encoder with positional encoding, followed by classification head for 4 affective states prediction.}
    \label{fig:transformer-architecture}
\end{figure*}

The proposed multimodal fusion method uses a hierarchical transformer-based architecture with three components: modality-specific encoding, additive fusion, and shared transformer encoder. Each input modality is processed by a dedicated transformer encoder branch, learning representations tailored to their unique characteristics. An additive fusion mechanism combines these representations, followed by a shared transformer encoder for further processing.

Let $X^{(1)}, X^{(2)}, \ldots, X^{(M)}$ denote input tensors for $M$ modalities, where $X^{(i)} \in \mathbb{R}^{N \times D_i}$ represents the $i$-th modality with $N$ samples and $D_i$ features. Each modality-specific encoder $f_i$ processes input $X^{(i)}$ to produce representation $Z^{(i)}$:
\begin{equation}
Z^{(i)} = f_i(X^{(i)})
\end{equation}

Additive fusion combines modality-specific representations:

\begin{equation}
Z_{\text{fused}} = \sum_{i=1}^{M} Z^{(i)}
\end{equation}

A shared transformer encoder $g$ processes $Z_{\text{fused}}$ to produce final encoded representation $H$:

\begin{equation}
H = g(Z_{\text{fused}})
\end{equation}

$H$ undergoes positional encoding and is fed into a classification head (linear layer) for affective state recognition.

This method captures interactions between diverse modalities while leveraging the transformer architecture's ability to model temporal dependencies and complex relationships within the fused representation~\cite{liang2022foundations}.

\subsubsection{Hyper-parameters}

The model's hyperparameters were selected using a gridsearch algorithm, testing various combinations on the full dataset. For each hyperparameter, we explored four different options, with numerical parameters varied by factors of 10. This search revealed that the model's performance was most sensitive to the learning rate and the choice of optimizer, while other parameters exhibited robust performance across a range of values.

The final hyperparameter configuration for the transformer model was as follows:

\begin{multicols}{2}
\begin{itemize}[noitemsep,topsep=0pt,parsep=0pt,partopsep=0pt,leftmargin=*]
    \item Transformer networks: \textit{1}
    \item Batch size: \textit{1024}
    \item Learning rate: \textit{0.0001}
    \item Number of epochs: \textit{50}
    \item Attention heads: \textit{2}
    \item Optimizer: \textit{RMSprop}
    \end{itemize}

\end{multicols}

We used k-fold (k = 29), training on 28 groups, and testing on the remaining one, resulting in 29 total evaluations. This approach was applied across all models, with multiple runs using different random seeds for robustness.

To mitigate overfitting, we implemented early stopping during the training process with a patience value of 5. This approach ensured that the model's training was halted when no improvement was observed in the validation loss for five consecutive epochs, thereby optimizing the model's generalization.

\section{Results}

We evaluated our proposed transformer model using various input modality configurations. Table~\ref{tab:ablation} presents an ablation study, showing F1 scores for different combinations of input modalities.

\begin{table}[!t]
\caption{Performance Comparison: Modality Combinations for Affective State Prediction}
\label{tab:ablation}
\centering
\small
\setlength{\tabcolsep}{4pt}
\begin{tabular}{@{}l S[table-format=2.0]@{}}
\toprule
Configuration & {F1 Score (\%)} \\
\midrule
GPT-4(V) Only & 23 \\
Full Context (FC) & 25 \\
Thermal Data & 30 \\
Thermal + FC & 58 \\
Action Units (AU) & 65 \\
AU + FC & 75 \\
Thermal + AU + GOC & 76 \\
Thermal + AU & 84 \\
Thermal + AU + FC & \textbf{89} \\
\bottomrule
\end{tabular}
\end{table}

\subsection{Thermal + Action Units + Full Context}

\begin{figure}
    \centering
    \includegraphics[width=0.8\linewidth]{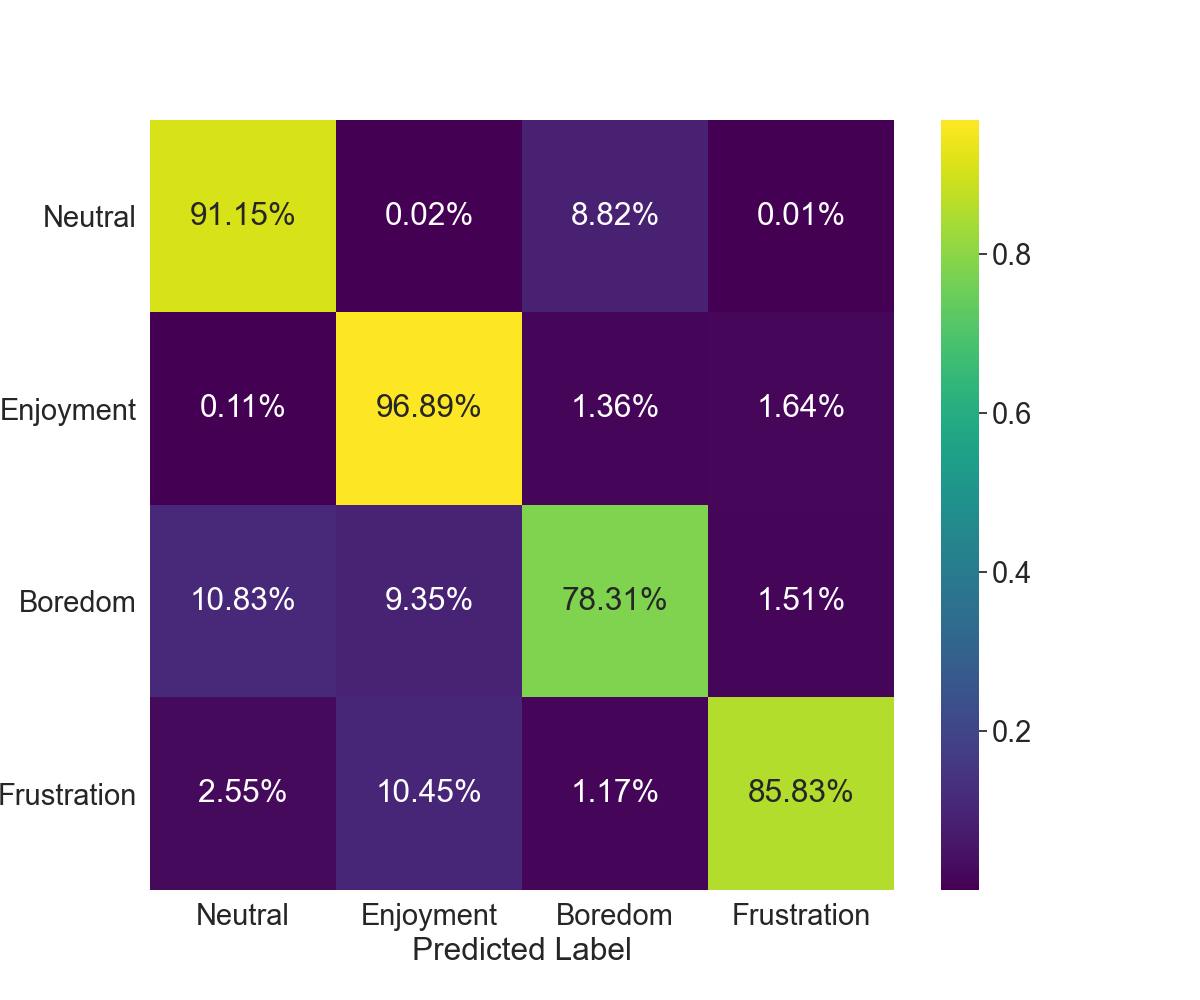}
    \caption{Confusion Matrix for Thermal + AU + FC configuration.}
    \label{fig:AUThermalContextConf}
\end{figure}

The combination of Thermal, Action Units (AU), and Full Context (FC) modalities yielded an F1 score of 89\%. Fig.~\ref{fig:AUThermalContextConf} presents the normalized confusion matrix for this configuration.

The model detected the neutral (baseline) state with 91.1\% accuracy. Enjoyment recognition achieved the highest accuracy at 96.9\%. Boredom detection showed an accuracy of 78.3\%, which is lower compared to other emotional states. Frustration detection reached an accuracy of 85.8\%.

Misclassifications were relatively low across categories. Notably, 10.83\% of boredom instances were misclassified as neutral, and 10.45\% of frustration instances were misclassified as enjoyment. Other misclassification rates remained below 10\%.

\subsection{Thermal + Action Units + Game-Only Context}

Excluding facial descriptions while retaining thermal data, action units, and game-only context resulted in an F1 score of 76\% and an average accuracy of 79.53\%. Baseline and enjoyment states were detected with high accuracy (84\% and 89\%, respectively), while boredom exhibited some confusion with enjoyment and baseline states.

\subsection{Thermal + Full Context}

The Thermal + FC configuration achieved an F1 score of 57.51\% and an average accuracy of 63.78\% (see Figure~\ref{fig:AUThermalConf}. The confusion matrix for this setup is as follows:

This configuration showed moderate performance, with the baseline state being the most accurately detected. However, there was notable confusion between other affective states, particularly between boredom and enjoyment, and between frustration and baseline.

\subsection{Individual Modalities}

\subsubsection{GPT-4V}
We have also used GPT-4V on each frame of the video with the prompt "Given the facial expressions and the context of playing a pacman game. Detect one output of four emotional states of the person in the image: Baseline, Boredom, Frustration and Enjoyment. Only write the output"
This resulted in an f1 score of 23\%, with major misclassification between all classes.

\subsubsection{Full Context Only}
Using only FC resulted in a low F1 score of 25.76\% and an average accuracy of 32.14\%, indicating significant misclassification across all affective states.

\subsubsection{Action Units Only}
The AU-only configuration achieved an F1 score of 64.85\% and an average accuracy of 67.38\%. It showed high accuracy for the baseline state (93.34\%) but exhibited confusion between frustration and enjoyment states.

\section{Discussion}

The results of our study underscore the importance of multimodal input configurations for affective state detection. The proposed transformer model demonstrated an improved performance when combining thermal data, action units (AU), and contextual information, achieving an F1 score of 89\%. This finding aligns with previous studies \cite{busso2004analysis, gunes2010automatic} that emphasize the effectiveness of multimodal approaches in affect recognition.

The configuration using thermal data and AU alone yielded an F1 score of 84\%, which is a substantial improvement over the use of either modality independently (30\% for thermal data alone and 65\% for AU alone). This result corroborates earlier research by \cite{pavlidis2002seeing} and \cite{ekman1978facial} on the value of thermal imagery and facial action units in emotion recognition. However, our results indicate that the combination of these modalities is more effective than using them in isolation, supporting the hypothesis that thermal and facial action data capture distinct but complementary aspects of affective expressions \cite{mohamed2023multi, mohamed2022automatic}.

Incorporating contextual information further enhanced the model's performance, as evidenced by the increase in the F1 score to 89\%. This improvement is particularly noteworthy in the recognition of enjoyment and frustration states, which showed significant gains in detection accuracy. This finding is supported by \cite{poria2017review}, who demonstrated that context-aware models could significantly enhance emotion recognition by providing additional situational cues that help disambiguate similar affective states.

Interestingly, the addition of Game Only-Context (GOC) led to a decreased F1 score compared to using only thermal and AU modalities, suggesting that GOC may introduce noise rather than providing sufficient context for the transformer. In contrast, the addition of Full-Context (FC) improved the accuracy of thermal data from 30\% to 58\% and AU data from 65\% to 75\%. These findings align with previous research by \cite{baltruvsaitis2018multimodal} and \cite{bosch2015multimodal}, emphasizing that the quality of added modalities is crucial, not just their quantity.

The reduction in F1 score when incorporating certain additional modalities is not unprecedented. \cite{5871582} observed similar effects in multimodal affect recognition, attributing such decreases to potential inter-modality conflicts or insufficient integration strategies.

Unlike the results in \cite{mohamed2023multi}, our transformer-based model with FC data (Figure \ref{fig:AUThermalContextConf}) nearly eliminates confusion between enjoyment and frustration, dramatically improving their classification (95.15\% and 85.82\% respectively).

The FC-only configuration performed poorly, with an F1 score of 25.76\%, indicating that contextual information alone is insufficient for accurate affective state detection. This is consistent with the findings of \cite{busso2004analysis}, who reported that while context can enhance emotion recognition, it cannot replace direct physiological or facial cues. Similarly, the single-modality configurations (thermal data or AU alone) showed limitations, particularly in differentiating between enjoyment and frustration or boredom and other states. These results underscore the necessity of multimodal approaches for affective state detection, as single modalities lack the comprehensive coverage needed to capture the full spectrum of affective state expressions.

Despite using a 7-second analysis window, our system is designed for real-time implementation on robots. This approach is effective because the system's primary goal is to detect the user's affective state during specific tasks, which typically takes 5-15 seconds.

\section{Conclusion}

Our study demonstrates the efficacy of multimodal integration for affective state detection. By fusing thermal data, action units, and contextual information, our transformer-based model achieved an impressive F1 score of 89\%, outperforming GPT-4(V)'s 23\%. This performance difference can be attributed to two key factors: First, our multimodal approach provides a more comprehensive view of affective states, capturing nuances that may be missed in purely visual analysis. Second, unlike GPT-4(V)'s general-purpose design, our model is specifically tailored for affective state detection in our experimental context.

These findings underscore the importance of diverse data sources and advanced fusion techniques in developing accurate affective state detection systems. Moreover, they show the potential for more advancements in affective computing through targeted multimodal approaches and specialized model architectures.

\bibliographystyle{acm}
\bibliography{bibliography}

\begin{thebibliography}{10}

\bibitem{baltruvsaitis2018challenges}
{\sc Baltru{\v{s}}aitis, T., Ahuja, C., and Morency, L.-P.}
\newblock Challenges and applications in multimodal machine learning.
\newblock {\em The Handbook of Multimodal-Multisensor Interfaces: Signal Processing, Architectures, and Detection of Emotion and Cognition-Volume 2\/} (2018), 17--48.

\bibitem{baltruvsaitis2018multimodal}
{\sc Baltru{\v{s}}aitis, T., Ahuja, C., and Morency, L.-P.}
\newblock Multimodal machine learning: A survey and taxonomy.
\newblock {\em IEEE transactions on pattern analysis and machine intelligence 41}, 2 (2018), 423--443.

\bibitem{bano2018multimodal}
{\sc Bano, S., Suveges, T., Zhang, J., and Mckenna, S.~J.}
\newblock Multimodal egocentric analysis of focused interactions.
\newblock {\em IEEE Access 6\/} (2018), 37493--37505.

\bibitem{barrett2019emotional}
{\sc Barrett, L.~F., Adolphs, R., Marsella, S., Martinez, A.~M., and Pollak, S.~D.}
\newblock Emotional expressions reconsidered: Challenges to inferring emotion from human facial movements.
\newblock {\em Psychological science in the public interest 20}, 1 (2019), 1--68.

\bibitem{bosch2015multimodal}
{\sc Bosch, N.}
\newblock Multimodal affect detection in the wild: Accuracy, availability, and generalizability.
\newblock In {\em Proceedings of the 2015 ACM on International Conference on Multimodal Interaction\/} (2015), pp.~645--649.

\bibitem{busso2004analysis}
{\sc Busso, C., Deng, Z., Yildirim, S., Bulut, M., Lee, C.~M., Kazemzadeh, A., Lee, S., Neumann, U., and Narayanan, S.}
\newblock Analysis of emotion recognition using facial expressions, speech and multimodal information.
\newblock In {\em Proceedings of the 6th International Conference on Multimodal Interfaces (ICMI)\/} (2004), pp.~205--211.

\bibitem{ekman1978facial}
{\sc Ekman, P., and Friesen, W.~V.}
\newblock Facial action coding system.
\newblock {\em Environmental Psychology \& Nonverbal Behavior\/} (1978).

\bibitem{faye2024context}
{\sc Faye, B., Azzag, H., Lebbah, M., and Bouchaffra, D.}
\newblock Context-based multimodal fusion.
\newblock {\em arXiv preprint arXiv:2403.04650\/} (2024).

\bibitem{gunes2010automatic}
{\sc Gunes, H., and Schuller, B.}
\newblock Automatic, dimensional and continuous emotion recognition.
\newblock {\em International Journal of Synthetic Emotions (IJSE) 1}, 1 (2010), 68--99.

\bibitem{pacmandesign}
{\sc Guneysu~Ozgur, A., Wessel, M.~J., Johal, W., Sharma, K., \"{O}zg\"{u}r, A., Vuadens, P., Mondada, F., Hummel, F.~C., and Dillenbourg, P.}
\newblock Iterative design of an upper limb rehabilitation game with tangible robots.
\newblock In {\em Proceedings of the 2018 ACM/IEEE International Conference on Human-Robot Interaction\/} (New York, NY, USA, 2018), HRI '18, Association for Computing Machinery, p.~241–250.

\bibitem{hoang2021context}
{\sc Hoang, M.-H., Kim, S.-H., Yang, H.-J., and Lee, G.-S.}
\newblock Context-aware emotion recognition based on visual relationship detection.
\newblock {\em IEEE Access 9\/} (2021), 90465--90474.

\bibitem{jiang2023understanding}
{\sc Jiang, Q., Chen, C., Zhao, H., Chen, L., Ping, Q., Tran, S.~D., Xu, Y., Zeng, B., and Chilimbi, T.}
\newblock Understanding and constructing latent modality structures in multi-modal representation learning.
\newblock {\em arXiv preprint arXiv:2303.05952\/} (2023).

\bibitem{kim2016emotion}
{\sc Kim, Y., Lee, H., and Kim, B.}
\newblock Emotion in context: Deep semantic feature fusion for video emotion recognition.
\newblock In {\em Proceedings of the 2016 ACM on Multimedia Conference\/} (2016), ACM, pp.~1275--1284.

\bibitem{lemaignan2024social}
{\sc Lemaignan, S., Andriella, A., Ferrini, L., Juricic, L., Mohamed, Y., and Ros, R.}
\newblock Social embeddings: Concept and initial investigation.
\newblock {\em Open Research Europe 4}, 63 (2024), 63.

\bibitem{liang2022foundations}
{\sc Liang, P.~P., Zadeh, A., and Morency, L.-P.}
\newblock Foundations and trends in multimodal machine learning: Principles, challenges, and open questions.
\newblock {\em arXiv preprint arXiv:2209.03430\/} (2022).

\bibitem{lu2024gpt}
{\sc Lu, H., Niu, X., Wang, J., Wang, Y., Hu, Q., Tang, J., Zhang, Y., Yuan, K., Huang, B., Yu, Z., et~al.}
\newblock Gpt as psychologist? preliminary evaluations for gpt-4v on visual affective computing.
\newblock {\em arXiv preprint arXiv:2403.05916\/} (2024).

\bibitem{ma2021facial}
{\sc Ma, F., Sun, B., and Li, S.}
\newblock Facial expression recognition with visual transformers and attentional selective fusion.
\newblock {\em IEEE Transactions on Affective Computing 14}, 2 (2021), 1236--1248.

\bibitem{mittal2021multimodal}
{\sc Mittal, T., Bera, A., and Manocha, D.}
\newblock Multimodal and context-aware emotion perception model with multiplicative fusion.
\newblock {\em IEEE MultiMedia 28}, 2 (2021), 67--75.

\bibitem{mohamed2022automatic}
{\sc Mohamed, Y., Ballardini, G., Parreira, M.~T., Lemaignan, S., and Leite, I.}
\newblock Automatic frustration detection using thermal imaging.
\newblock In {\em 2022 17th ACM/IEEE International Conference on Human-Robot Interaction (HRI)\/} (2022), IEEE, pp.~451--459.

\bibitem{mohamed2023multi}
{\sc Mohamed, Y., G{\"u}neysu, A., Lemaignan, S., and Leite, I.}
\newblock Multi-modal affect detection using thermal and optical imaging in a gamified robotic exercise.
\newblock {\em International Journal of Social Robotics\/} (2023), 1--17.

\bibitem{openaiembeddings}
{\sc OpenAI}.
\newblock Openai embeddings: Language models as apis.
\newblock {\em OpenAI Blog\/} (March 2021).

\bibitem{ekman2005}
{\sc Paul, E., and Rosenberg, E.~E.}
\newblock {\em What the face reveals : basic and applied studies of spontaneous expression using the facial action coding system (FACS)}.
\newblock Oxford University Press, 2005.

\bibitem{pavlidis2002seeing}
{\sc Pavlidis, I., Eberhardt, N.~L., and Levine, J.~A.}
\newblock Seeing through the face of deception.
\newblock {\em Nature 415}, 6867 (2002), 35--35.

\bibitem{poria2017review}
{\sc Poria, S., Cambria, E., Bajpai, R., and Hussain, A.}
\newblock A review of affective computing: From unimodal analysis to multimodal fusion.
\newblock {\em Information Fusion 37\/} (2017), 98--125.

\bibitem{vaswani2017attention}
{\sc Vaswani, A., Shazeer, N., Parmar, N., Uszkoreit, J., Jones, L., Gomez, A.~N., Kaiser, {\L}., and Polosukhin, I.}
\newblock Attention is all you need.
\newblock {\em Advances in neural information processing systems 30\/} (2017).

\bibitem{5871582}
{\sc Wagner, J., Andre, E., Lingenfelser, F., and Kim, J.}
\newblock Exploring fusion methods for multimodal emotion recognition with missing data.
\newblock {\em IEEE Transactions on Affective Computing 2}, 4 (2011), 206--218.

\bibitem{wang2024exploiting}
{\sc Wang, J., Liu, B., Miao, C., Zhao, Z., Zhuang, W., Chu, Q., and Yu, N.}
\newblock Exploiting modality-specific features for multi-modal manipulation detection and grounding.
\newblock In {\em ICASSP 2024-2024 IEEE International Conference on Acoustics, Speech and Signal Processing (ICASSP)\/} (2024), IEEE, pp.~4935--4939.

\bibitem{wang2019context}
{\sc Wang, Y., Li, M., Liu, S., and Li, M.}
\newblock Context-aware emotion recognition networks.
\newblock {\em arXiv preprint arXiv:1908.05913\/} (2019).

\end{thebibliography}

\end{document}